\newcounter{treecount}
\newcounter{branchcount}
\newsavebox{\parentbox}
\newsavebox{\treebox}
\newsavebox{\treeboxone}
\newsavebox{\treeboxtwo}
\newsavebox{\treeboxthree}
\newsavebox{\treeboxfour}
\newsavebox{\treeboxfive}
\newsavebox{\treeboxsix}
\newsavebox{\treeboxseven}
\newsavebox{\treeboxeight}
\newsavebox{\treeboxnine}
\newsavebox{\treeboxten}
\newsavebox{\treeboxeleven}
\newsavebox{\treeboxtwelve}
\newsavebox{\treeboxthirteen}
\newsavebox{\treeboxfourteen}
\newsavebox{\treeboxfifteen}
\newsavebox{\treeboxsixteen}
\newsavebox{\treeboxseventeen}
\newsavebox{\treeboxeighteen}
\newsavebox{\treeboxnineteen}
\newsavebox{\treeboxtwenty}
\newlength{\treeoffsetone}
\newlength{\treeoffsettwo}
\newlength{\treeoffsetthree}
\newlength{\treeoffsetfour}
\newlength{\treeoffsetfive}
\newlength{\treeoffsetsix}
\newlength{\treeoffsetseven}
\newlength{\treeoffseteight}
\newlength{\treeoffsetnine}
\newlength{\treeoffsetten}
\newlength{\treeoffseteleven}
\newlength{\treeoffsettwelve}
\newlength{\treeoffsetthirteen}
\newlength{\treeoffsetfourteen}
\newlength{\treeoffsetfifteen}
\newlength{\treeoffsetsixteen}
\newlength{\treeoffsetseventeen}
\newlength{\treeoffseteighteen}
\newlength{\treeoffsetnineteen}
\newlength{\treeoffsettwenty}

\newlength{\treeshiftone}
\newlength{\treeshifttwo}
\newlength{\treeshiftthree}
\newlength{\treeshiftfour}
\newlength{\treeshiftfive}
\newlength{\treeshiftsix}
\newlength{\treeshiftseven}
\newlength{\treeshifteight}
\newlength{\treeshiftnine}
\newlength{\treeshiftten}
\newlength{\treeshifteleven}
\newlength{\treeshifttwelve}
\newlength{\treeshiftthirteen}
\newlength{\treeshiftfourteen}
\newlength{\treeshiftfifteen}
\newlength{\treeshiftsixteen}
\newlength{\treeshiftseventeen}
\newlength{\treeshifteighteen}
\newlength{\treeshiftnineteen}
\newlength{\treeshifttwenty}
\newlength{\treewidthone}
\newlength{\treewidthtwo}
\newlength{\treewidththree}
\newlength{\treewidthfour}
\newlength{\treewidthfive}
\newlength{\treewidthsix}
\newlength{\treewidthseven}
\newlength{\treewidtheight}
\newlength{\treewidthnine}
\newlength{\treewidthten}
\newlength{\treewidtheleven}
\newlength{\treewidthtwelve}
\newlength{\treewidththirteen}
\newlength{\treewidthfourteen}
\newlength{\treewidthfifteen}
\newlength{\treewidthsixteen}
\newlength{\treewidthseventeen}
\newlength{\treewidtheighteen}
\newlength{\treewidthnineteen}
\newlength{\treewidthtwenty}
\newlength{\daughteroffsetone}
\newlength{\daughteroffsettwo}
\newlength{\daughteroffsetthree}
\newlength{\daughteroffsetfour}
\newlength{\branchwidthone}
\newlength{\branchwidthtwo}
\newlength{\branchwidththree}
\newlength{\branchwidthfour}
\newlength{\parentoffset}
\newlength{\treeoffset}
\newlength{\daughteroffset}
\newlength{\branchwidth}
\newlength{\parentwidth}
\newlength{\treewidth}
\newcommand{\ontop}[1]{\begin{tabular}{c}#1\end{tabular}}
\newcommand{\poptree}{%
\ifnum\value{treecount}=0\typeout{QobiTeX warning---Tree stack underflow}\fi%
\addtocounter{treecount}{-1}%
\setlength{\treeoffsettwo}{\treeoffsetthree}%
\setlength{\treeoffsetthree}{\treeoffsetfour}%
\setlength{\treeoffsetfour}{\treeoffsetfive}%
\setlength{\treeoffsetfive}{\treeoffsetsix}%
\setlength{\treeoffsetsix}{\treeoffsetseven}%
\setlength{\treeoffsetseven}{\treeoffseteight}%
\setlength{\treeoffseteight}{\treeoffsetnine}%
\setlength{\treeoffsetnine}{\treeoffsetten}%
\setlength{\treeoffsetten}{\treeoffseteleven}%
\setlength{\treeoffseteleven}{\treeoffsettwelve}%
\setlength{\treeoffsettwelve}{\treeoffsetthirteen}%
\setlength{\treeoffsetthirteen}{\treeoffsetfourteen}%
\setlength{\treeoffsetfourteen}{\treeoffsetfifteen}%
\setlength{\treeoffsetfifteen}{\treeoffsetsixteen}%
\setlength{\treeoffsetsixteen}{\treeoffsetseventeen}%
\setlength{\treeoffsetseventeen}{\treeoffseteighteen}%
\setlength{\treeoffseteighteen}{\treeoffsetnineteen}%
\setlength{\treeoffsetnineteen}{\treeoffsettwenty}%
\setlength{\treeshifttwo}{\treeshiftthree}%
\setlength{\treeshiftthree}{\treeshiftfour}%
\setlength{\treeshiftfour}{\treeshiftfive}%
\setlength{\treeshiftfive}{\treeshiftsix}%
\setlength{\treeshiftsix}{\treeshiftseven}%
\setlength{\treeshiftseven}{\treeshifteight}%
\setlength{\treeshifteight}{\treeshiftnine}%
\setlength{\treeshiftnine}{\treeshiftten}%
\setlength{\treeshiftten}{\treeshifteleven}%
\setlength{\treeshifteleven}{\treeshifttwelve}%
\setlength{\treeshifttwelve}{\treeshiftthirteen}%
\setlength{\treeshiftthirteen}{\treeshiftfourteen}%
\setlength{\treeshiftfourteen}{\treeshiftfifteen}%
\setlength{\treeshiftfifteen}{\treeshiftsixteen}%
\setlength{\treeshiftsixteen}{\treeshiftseventeen}%
\setlength{\treeshiftseventeen}{\treeshifteighteen}%
\setlength{\treeshifteighteen}{\treeshiftnineteen}%
\setlength{\treeshiftnineteen}{\treeshifttwenty}%
\setlength{\treewidthtwo}{\treewidththree}%
\setlength{\treewidththree}{\treewidthfour}%
\setlength{\treewidthfour}{\treewidthfive}%
\setlength{\treewidthfive}{\treewidthsix}%
\setlength{\treewidthsix}{\treewidthseven}%
\setlength{\treewidthseven}{\treewidtheight}%
\setlength{\treewidtheight}{\treewidthnine}%
\setlength{\treewidthnine}{\treewidthten}%
\setlength{\treewidthten}{\treewidtheleven}%
\setlength{\treewidtheleven}{\treewidthtwelve}%
\setlength{\treewidthtwelve}{\treewidththirteen}%
\setlength{\treewidththirteen}{\treewidthfourteen}%
\setlength{\treewidthfourteen}{\treewidthfifteen}%
\setlength{\treewidthfifteen}{\treewidthsixteen}%
\setlength{\treewidthsixteen}{\treewidthseventeen}%
\setlength{\treewidthseventeen}{\treewidtheighteen}%
\setlength{\treewidtheighteen}{\treewidthnineteen}%
\setlength{\treewidthnineteen}{\treewidthtwenty}%
\sbox{\treeboxtwo}{\usebox{\treeboxthree}}%
\sbox{\treeboxthree}{\usebox{\treeboxfour}}%
\sbox{\treeboxfour}{\usebox{\treeboxfive}}%
\sbox{\treeboxfive}{\usebox{\treeboxsix}}%
\sbox{\treeboxsix}{\usebox{\treeboxseven}}%
\sbox{\treeboxseven}{\usebox{\treeboxeight}}%
\sbox{\treeboxeight}{\usebox{\treeboxnine}}%
\sbox{\treeboxnine}{\usebox{\treeboxten}}%
\sbox{\treeboxten}{\usebox{\treeboxeleven}}%
\sbox{\treeboxeleven}{\usebox{\treeboxtwelve}}%
\sbox{\treeboxtwelve}{\usebox{\treeboxthirteen}}%
\sbox{\treeboxthirteen}{\usebox{\treeboxfourteen}}%
\sbox{\treeboxfourteen}{\usebox{\treeboxfifteen}}%
\sbox{\treeboxfifteen}{\usebox{\treeboxsixteen}}%
\sbox{\treeboxsixteen}{\usebox{\treeboxseventeen}}%
\sbox{\treeboxseventeen}{\usebox{\treeboxeighteen}}%
\sbox{\treeboxeighteen}{\usebox{\treeboxnineteen}}%
\sbox{\treeboxnineteen}{\usebox{\treeboxtwenty}}}
\newcommand{\leaf}[1]{%
\ifnum\value{treecount}=20\typeout{QobiTeX warning---Tree stack overflow}\fi%
\addtocounter{treecount}{1}%
\sbox{\treeboxtwenty}{\usebox{\treeboxnineteen}}%
\sbox{\treeboxnineteen}{\usebox{\treeboxeighteen}}%
\sbox{\treeboxeighteen}{\usebox{\treeboxseventeen}}%
\sbox{\treeboxseventeen}{\usebox{\treeboxsixteen}}%
\sbox{\treeboxsixteen}{\usebox{\treeboxfifteen}}%
\sbox{\treeboxfifteen}{\usebox{\treeboxfourteen}}%
\sbox{\treeboxfourteen}{\usebox{\treeboxthirteen}}%
\sbox{\treeboxthirteen}{\usebox{\treeboxtwelve}}%
\sbox{\treeboxtwelve}{\usebox{\treeboxeleven}}%
\sbox{\treeboxeleven}{\usebox{\treeboxten}}%
\sbox{\treeboxten}{\usebox{\treeboxnine}}%
\sbox{\treeboxnine}{\usebox{\treeboxeight}}%
\sbox{\treeboxeight}{\usebox{\treeboxseven}}%
\sbox{\treeboxseven}{\usebox{\treeboxsix}}%
\sbox{\treeboxsix}{\usebox{\treeboxfive}}%
\sbox{\treeboxfive}{\usebox{\treeboxfour}}%
\sbox{\treeboxfour}{\usebox{\treeboxthree}}%
\sbox{\treeboxthree}{\usebox{\treeboxtwo}}%
\sbox{\treeboxtwo}{\usebox{\treeboxone}}%
\sbox{\treeboxone}{\ontop{#1}}%
\sbox{\treeboxone}{\raisebox{-\ht\treeboxone}{\usebox{\treeboxone}}}%
\setlength{\treeoffsettwenty}{\treeoffsetnineteen}%
\setlength{\treeoffsetnineteen}{\treeoffseteighteen}%
\setlength{\treeoffseteighteen}{\treeoffsetseventeen}%
\setlength{\treeoffsetseventeen}{\treeoffsetsixteen}%
\setlength{\treeoffsetsixteen}{\treeoffsetfifteen}%
\setlength{\treeoffsetfifteen}{\treeoffsetfourteen}%
\setlength{\treeoffsetfourteen}{\treeoffsetthirteen}%
\setlength{\treeoffsetthirteen}{\treeoffsettwelve}%
\setlength{\treeoffsettwelve}{\treeoffseteleven}%
\setlength{\treeoffseteleven}{\treeoffsetten}%
\setlength{\treeoffsetten}{\treeoffsetnine}%
\setlength{\treeoffsetnine}{\treeoffseteight}%
\setlength{\treeoffseteight}{\treeoffsetseven}%
\setlength{\treeoffsetseven}{\treeoffsetsix}%
\setlength{\treeoffsetsix}{\treeoffsetfive}%
\setlength{\treeoffsetfive}{\treeoffsetfour}%
\setlength{\treeoffsetfour}{\treeoffsetthree}%
\setlength{\treeoffsetthree}{\treeoffsettwo}%
\setlength{\treeoffsettwo}{\treeoffsetone}%
\setlength{\treeoffsetone}{0.5\wd\treeboxone}%
\setlength{\treeshifttwenty}{\treeshiftnineteen}%
\setlength{\treeshiftnineteen}{\treeshifteighteen}%
\setlength{\treeshifteighteen}{\treeshiftseventeen}%
\setlength{\treeshiftseventeen}{\treeshiftsixteen}%
\setlength{\treeshiftsixteen}{\treeshiftfifteen}%
\setlength{\treeshiftfifteen}{\treeshiftfourteen}%
\setlength{\treeshiftfourteen}{\treeshiftthirteen}%
\setlength{\treeshiftthirteen}{\treeshifttwelve}%
\setlength{\treeshifttwelve}{\treeshifteleven}%
\setlength{\treeshifteleven}{\treeshiftten}%
\setlength{\treeshiftten}{\treeshiftnine}%
\setlength{\treeshiftnine}{\treeshifteight}%
\setlength{\treeshifteight}{\treeshiftseven}%
\setlength{\treeshiftseven}{\treeshiftsix}%
\setlength{\treeshiftsix}{\treeshiftfive}%
\setlength{\treeshiftfive}{\treeshiftfour}%
\setlength{\treeshiftfour}{\treeshiftthree}%
\setlength{\treeshiftthree}{\treeshifttwo}%
\setlength{\treeshifttwo}{\treeshiftone}%
\setlength{\treeshiftone}{0pt}%
\setlength{\treewidthtwenty}{\treewidthnineteen}%
\setlength{\treewidthnineteen}{\treewidtheighteen}%
\setlength{\treewidtheighteen}{\treewidthseventeen}%
\setlength{\treewidthseventeen}{\treewidthsixteen}%
\setlength{\treewidthsixteen}{\treewidthfifteen}%
\setlength{\treewidthfifteen}{\treewidthfourteen}%
\setlength{\treewidthfourteen}{\treewidththirteen}%
\setlength{\treewidththirteen}{\treewidthtwelve}%
\setlength{\treewidthtwelve}{\treewidtheleven}%
\setlength{\treewidtheleven}{\treewidthten}%
\setlength{\treewidthten}{\treewidthnine}%
\setlength{\treewidthnine}{\treewidtheight}%
\setlength{\treewidtheight}{\treewidthseven}%
\setlength{\treewidthseven}{\treewidthsix}%
\setlength{\treewidthsix}{\treewidthfive}%
\setlength{\treewidthfive}{\treewidthfour}%
\setlength{\treewidthfour}{\treewidththree}%
\setlength{\treewidththree}{\treewidthtwo}%
\setlength{\treewidthtwo}{\treewidthone}%
\setlength{\treewidthone}{\wd\treeboxone}}
\newcommand{\branch}[2]{%
\setcounter{branchcount}{#1}%
\ifnum\value{branchcount}=1\sbox{\parentbox}{\ontop{#2}}%
\setlength{\parentoffset}{\treeoffsetone}%
\addtolength{\parentoffset}{-0.5\wd\parentbox}%
\setlength{\daughteroffset}{0in}%
\ifdim\parentoffset<0in%
\setlength{\daughteroffset}{-\parentoffset}%
\setlength{\parentoffset}{0in}\fi%
\setlength{\parentwidth}{\parentoffset}%
\addtolength{\parentwidth}{\wd\parentbox}%
\setlength{\treeoffset}{\daughteroffset}%
\addtolength{\treeoffset}{\treeoffsetone}%
\setlength{\treewidth}{\wd\treeboxone}%
\addtolength{\treewidth}{\daughteroffset}%
\ifdim\treewidth<\parentwidth\setlength{\treewidth}{\parentwidth}\fi%
\sbox{\treebox}{\begin{minipage}{\treewidth}%
\begin{flushleft}%
\hspace*{\parentoffset}\usebox{\parentbox}\\
{\setlength{\unitlength}{2ex}%
\hspace*{\treeoffset}\begin{picture}(0,1)%
\put(0,0){\line(0,1){1}}%
\end{picture}}\\
\vspace{-\baselineskip}
\hspace*{\daughteroffset}%
\raisebox{-\ht\treeboxone}{\usebox{\treeboxone}}%
\end{flushleft}%
\end{minipage}}%
\setlength{\treeoffsetone}{\parentoffset}%
\addtolength{\treeoffsetone}{0.5\wd\parentbox}%
\setlength{\treeshiftone}{0pt}%
\setlength{\treewidthone}{\treewidth}%
\sbox{\treeboxone}{\usebox{\treebox}}%
\else\ifnum\value{branchcount}=2\sbox{\parentbox}{\ontop{#2}}%
\setlength{\branchwidthone}{\treewidthtwo}%
\addtolength{\branchwidthone}{\treeoffsetone}%
\addtolength{\branchwidthone}{-\treeshiftone}%
\addtolength{\branchwidthone}{-\treeoffsettwo}%
\setlength{\branchwidth}{\branchwidthone}%
\setlength{\daughteroffsetone}{\branchwidth}%
\addtolength{\daughteroffsetone}{-\branchwidthone}%
\addtolength{\daughteroffsetone}{-\treeshiftone}%
\setlength{\parentoffset}{-0.5\wd\parentbox}%
\addtolength{\parentoffset}{\treeoffsettwo}%
\addtolength{\parentoffset}{0.5\branchwidth}%
\setlength{\daughteroffset}{0in}%
\ifdim\parentoffset<0in%
\setlength{\daughteroffset}{-\parentoffset}%
\setlength{\parentoffset}{0in}\fi%
\setlength{\parentwidth}{\parentoffset}%
\addtolength{\parentwidth}{\wd\parentbox}%
\setlength{\treeoffset}{\daughteroffset}%
\addtolength{\treeoffset}{\treeoffsettwo}%
\setlength{\treewidth}{\wd\treeboxone}%
\addtolength{\treewidth}{\daughteroffsetone}%
\addtolength{\treewidth}{\treewidthtwo}%
\addtolength{\treewidth}{\daughteroffset}%
\ifdim\treewidth<\parentwidth\setlength{\treewidth}{\parentwidth}\fi%
\sbox{\treebox}{\begin{minipage}{\treewidth}%
\begin{flushleft}%
\hspace*{\parentoffset}\usebox{\parentbox}\\
{\setlength{\unitlength}{0.5\branchwidth}%
\hspace*{\treeoffset}\begin{picture}(2,0.5)%
\put(0,0){\line(2,1){1}}%
\put(2,0){\line(-2,1){1}}%
\end{picture}}\\
\vspace{-\baselineskip}
\hspace*{\daughteroffset}%
\makebox[\treewidthtwo][l]%
{\raisebox{-\ht\treeboxtwo}{\usebox{\treeboxtwo}}}%
\hspace*{\daughteroffsetone}%
\raisebox{-\ht\treeboxone}{\usebox{\treeboxone}}%
\end{flushleft}%
\end{minipage}}%
\setlength{\treeoffsetone}{\parentoffset}%
\addtolength{\treeoffsetone}{0.5\wd\parentbox}%
\setlength{\treeshiftone}{0pt}%
\setlength{\treewidthone}{\treewidth}%
\sbox{\treeboxone}{\usebox{\treebox}}\poptree%
\else\ifnum\value{branchcount}=3\sbox{\parentbox}{\ontop{#2}}%
\setlength{\branchwidthone}{\treewidthtwo}%
\addtolength{\branchwidthone}{\treeoffsetone}%
\addtolength{\branchwidthone}{-\treeshiftone}%
\addtolength{\branchwidthone}{-\treeoffsettwo}%
\setlength{\branchwidthtwo}{\treewidththree}%
\addtolength{\branchwidthtwo}{\treeoffsettwo}%
\addtolength{\branchwidthtwo}{-\treeshifttwo}%
\addtolength{\branchwidthtwo}{-\treeoffsetthree}%
\setlength{\branchwidth}{\branchwidthone}%
\ifdim\branchwidthtwo>\branchwidth%
\setlength{\branchwidth}{\branchwidthtwo}\fi%
\setlength{\daughteroffsetone}{\branchwidth}%
\addtolength{\daughteroffsetone}{-\branchwidthone}%
\addtolength{\daughteroffsetone}{-\treeshiftone}%
\setlength{\daughteroffsettwo}{\branchwidth}%
\addtolength{\daughteroffsettwo}{-\branchwidthtwo}%
\addtolength{\daughteroffsettwo}{-\treeshifttwo}%
\setlength{\parentoffset}{-0.5\wd\parentbox}%
\addtolength{\parentoffset}{\treeoffsetthree}%
\addtolength{\parentoffset}{\branchwidth}%
\setlength{\daughteroffset}{0in}%
\ifdim\parentoffset<0in%
\setlength{\daughteroffset}{-\parentoffset}%
\setlength{\parentoffset}{0in}\fi%
\setlength{\parentwidth}{\parentoffset}%
\addtolength{\parentwidth}{\wd\parentbox}%
\setlength{\treeoffset}{\daughteroffset}%
\addtolength{\treeoffset}{\treeoffsetthree}%
\setlength{\treewidth}{\wd\treeboxone}%
\addtolength{\treewidth}{\daughteroffsetone}%
\addtolength{\treewidth}{\treewidthtwo}%
\addtolength{\treewidth}{\daughteroffsettwo}%
\addtolength{\treewidth}{\treewidththree}%
\addtolength{\treewidth}{\daughteroffset}%
\ifdim\treewidth<\parentwidth\setlength{\treewidth}{\parentwidth}\fi%
\sbox{\treebox}{\begin{minipage}{\treewidth}%
\begin{flushleft}%
\hspace*{\parentoffset}\usebox{\parentbox}\\
{\setlength{\unitlength}{0.5\branchwidth}%
\hspace*{\treeoffset}\begin{picture}(4,1)%
\put(0,0){\line(2,1){2}}%
\put(2,0){\line(0,1){1}}%
\put(4,0){\line(-2,1){2}}%
\end{picture}}\\
\vspace{-\baselineskip}
\hspace*{\daughteroffset}%
\makebox[\treewidththree][l]%
{\raisebox{-\ht\treeboxthree}{\usebox{\treeboxthree}}}%
\hspace*{\daughteroffsettwo}%
\makebox[\treewidthtwo][l]%
{\raisebox{-\ht\treeboxtwo}{\usebox{\treeboxtwo}}}%
\hspace*{\daughteroffsetone}%
\raisebox{-\ht\treeboxone}{\usebox{\treeboxone}}%
\end{flushleft}%
\end{minipage}}%
\setlength{\treeoffsetone}{\parentoffset}%
\addtolength{\treeoffsetone}{0.5\wd\parentbox}%
\setlength{\treeshiftone}{0pt}%
\setlength{\treewidthone}{\treewidth}%
\sbox{\treeboxone}{\usebox{\treebox}}\poptree\poptree%
\else\ifnum\value{branchcount}=4\sbox{\parentbox}{\ontop{#2}}%
\setlength{\branchwidthone}{\treewidthtwo}%
\addtolength{\branchwidthone}{\treeoffsetone}%
\addtolength{\branchwidthone}{-\treeshiftone}%
\addtolength{\branchwidthone}{-\treeoffsettwo}%
\setlength{\branchwidthtwo}{\treewidththree}%
\addtolength{\branchwidthtwo}{\treeoffsettwo}%
\addtolength{\branchwidthtwo}{-\treeshifttwo}%
\addtolength{\branchwidthtwo}{-\treeoffsetthree}%
\setlength{\branchwidththree}{\treewidthfour}%
\addtolength{\branchwidththree}{\treeoffsetthree}%
\addtolength{\branchwidththree}{-\treeshiftthree}%
\addtolength{\branchwidththree}{-\treeoffsetfour}%
\setlength{\branchwidth}{\branchwidthone}%
\ifdim\branchwidthtwo>\branchwidth%
\setlength{\branchwidth}{\branchwidthtwo}\fi%
\ifdim\branchwidththree>\branchwidth%
\setlength{\branchwidth}{\branchwidththree}\fi%
\setlength{\daughteroffsetone}{\branchwidth}%
\addtolength{\daughteroffsetone}{-\branchwidthone}%
\addtolength{\daughteroffsetone}{-\treeshiftone}%
\setlength{\daughteroffsettwo}{\branchwidth}%
\addtolength{\daughteroffsettwo}{-\branchwidthtwo}%
\addtolength{\daughteroffsettwo}{-\treeshifttwo}%
\setlength{\daughteroffsetthree}{\branchwidth}%
\addtolength{\daughteroffsetthree}{-\branchwidththree}%
\addtolength{\daughteroffsetthree}{-\treeshiftthree}%
\setlength{\parentoffset}{-0.5\wd\parentbox}%
\addtolength{\parentoffset}{\treeoffsetfour}%
\addtolength{\parentoffset}{1.5\branchwidth}%
\setlength{\daughteroffset}{0in}%
\ifdim\parentoffset<0in%
\setlength{\daughteroffset}{-\parentoffset}%
\setlength{\parentoffset}{0in}\fi%
\setlength{\parentwidth}{\parentoffset}%
\addtolength{\parentwidth}{\wd\parentbox}%
\setlength{\treeoffset}{\daughteroffset}%
\addtolength{\treeoffset}{\treeoffsetfour}%
\setlength{\treewidth}{\wd\treeboxone}%
\addtolength{\treewidth}{\daughteroffsetone}%
\addtolength{\treewidth}{\treewidthtwo}%
\addtolength{\treewidth}{\daughteroffsettwo}%
\addtolength{\treewidth}{\treewidththree}%
\addtolength{\treewidth}{\daughteroffsetthree}%
\addtolength{\treewidth}{\treewidthfour}%
\addtolength{\treewidth}{\daughteroffset}%
\ifdim\treewidth<\parentwidth\setlength{\treewidth}{\parentwidth}\fi%
\sbox{\treebox}{\begin{minipage}{\treewidth}%
\begin{flushleft}%
\hspace*{\parentoffset}\usebox{\parentbox}\\
{\setlength{\unitlength}{0.5\branchwidth}%
\hspace*{\treeoffset}\begin{picture}(6,1)%
\put(0,0){\line(3,1){3}}%
\put(2,0){\line(1,1){1}}%
\put(4,0){\line(-1,1){1}}%
\put(6,0){\line(-3,1){3}}%
\end{picture}}\\
\vspace{-\baselineskip}
\hspace*{\daughteroffset}%
\makebox[\treewidthfour][l]%
{\raisebox{-\ht\treeboxfour}{\usebox{\treeboxfour}}}%
\hspace*{\daughteroffsetthree}%
\makebox[\treewidththree][l]%
{\raisebox{-\ht\treeboxthree}{\usebox{\treeboxthree}}}%
\hspace*{\daughteroffsettwo}%
\makebox[\treewidthtwo][l]%
{\raisebox{-\ht\treeboxtwo}{\usebox{\treeboxtwo}}}%
\hspace*{\daughteroffsetone}%
\raisebox{-\ht\treeboxone}{\usebox{\treeboxone}}%
\end{flushleft}%
\end{minipage}}%
\setlength{\treeoffsetone}{\parentoffset}%
\addtolength{\treeoffsetone}{0.5\wd\parentbox}%
\setlength{\treeshiftone}{0pt}%
\setlength{\treewidthone}{\treewidth}%
\sbox{\treeboxone}{\usebox{\treebox}}\poptree\poptree\poptree%
\else\ifnum\value{branchcount}=5\sbox{\parentbox}{\ontop{#2}}%
\setlength{\branchwidthone}{\treewidthtwo}%
\addtolength{\branchwidthone}{\treeoffsetone}%
\addtolength{\branchwidthone}{-\treeshiftone}%
\addtolength{\branchwidthone}{-\treeoffsettwo}%
\setlength{\branchwidthtwo}{\treewidththree}%
\addtolength{\branchwidthtwo}{\treeoffsettwo}%
\addtolength{\branchwidthtwo}{-\treeshifttwo}%
\addtolength{\branchwidthtwo}{-\treeoffsetthree}%
\setlength{\branchwidththree}{\treewidthfour}%
\addtolength{\branchwidththree}{\treeoffsetthree}%
\addtolength{\branchwidththree}{-\treeshiftthree}%
\addtolength{\branchwidththree}{-\treeoffsetfour}%
\setlength{\branchwidthfour}{\treewidthfive}%
\addtolength{\branchwidthfour}{\treeoffsetfour}%
\addtolength{\branchwidthfour}{-\treeshiftfour}%
\addtolength{\branchwidthfour}{-\treeoffsetfive}%
\setlength{\branchwidth}{\branchwidthone}%
\ifdim\branchwidthtwo>\branchwidth%
\setlength{\branchwidth}{\branchwidthtwo}\fi%
\ifdim\branchwidththree>\branchwidth%
\setlength{\branchwidth}{\branchwidththree}\fi%
\ifdim\branchwidthfour>\branchwidth%
\setlength{\branchwidth}{\branchwidthfour}\fi%
\setlength{\daughteroffsetone}{\branchwidth}%
\addtolength{\daughteroffsetone}{-\branchwidthone}%
\addtolength{\daughteroffsetone}{-\treeshiftone}%
\setlength{\daughteroffsettwo}{\branchwidth}%
\addtolength{\daughteroffsettwo}{-\branchwidthtwo}%
\addtolength{\daughteroffsettwo}{-\treeshifttwo}%
\setlength{\daughteroffsetthree}{\branchwidth}%
\addtolength{\daughteroffsetthree}{-\branchwidththree}%
\addtolength{\daughteroffsetthree}{-\treeshiftthree}%
\setlength{\daughteroffsetfour}{\branchwidth}%
\addtolength{\daughteroffsetfour}{-\branchwidthfour}%
\addtolength{\daughteroffsetfour}{-\treeshiftfour}%
\setlength{\parentoffset}{-0.5\wd\parentbox}%
\addtolength{\parentoffset}{\treeoffsetfive}%
\addtolength{\parentoffset}{2\branchwidth}%
\setlength{\daughteroffset}{0in}%
\ifdim\parentoffset<0in%
\setlength{\daughteroffset}{-\parentoffset}%
\setlength{\parentoffset}{0in}\fi%
\setlength{\parentwidth}{\parentoffset}%
\addtolength{\parentwidth}{\wd\parentbox}%
\setlength{\treeoffset}{\daughteroffset}%
\addtolength{\treeoffset}{\treeoffsetfive}%
\setlength{\treewidth}{\wd\treeboxone}%
\addtolength{\treewidth}{\daughteroffsetone}%
\addtolength{\treewidth}{\treewidthtwo}%
\addtolength{\treewidth}{\daughteroffsettwo}%
\addtolength{\treewidth}{\treewidththree}%
\addtolength{\treewidth}{\daughteroffsetthree}%
\addtolength{\treewidth}{\treewidthfour}%
\addtolength{\treewidth}{\daughteroffsetfour}%
\addtolength{\treewidth}{\treewidthfive}%
\addtolength{\treewidth}{\daughteroffset}%
\ifdim\treewidth<\parentwidth\setlength{\treewidth}{\parentwidth}\fi%
\sbox{\treebox}{\begin{minipage}{\treewidth}%
\begin{flushleft}%
\hspace*{\parentoffset}\usebox{\parentbox}\\
{\setlength{\unitlength}{0.5\branchwidth}%
\hspace*{\treeoffset}\begin{picture}(8,1)%
\put(0,0){\line(4,1){4}}%
\put(2,0){\line(2,1){2}}%
\put(4,0){\line(0,1){1}}%
\put(6,0){\line(-2,1){2}}%
\put(8,0){\line(-4,1){4}}%
\end{picture}}\\
\vspace{-\baselineskip}
\hspace*{\daughteroffset}%
\makebox[\treewidthfive][l]%
{\raisebox{-\ht\treeboxfour}{\usebox{\treeboxfive}}}%
\hspace*{\daughteroffsetfour}%
\makebox[\treewidthfour][l]%
{\raisebox{-\ht\treeboxfour}{\usebox{\treeboxfour}}}%
\hspace*{\daughteroffsetthree}%
\makebox[\treewidththree][l]%
{\raisebox{-\ht\treeboxthree}{\usebox{\treeboxthree}}}%
\hspace*{\daughteroffsettwo}%
\makebox[\treewidthtwo][l]%
{\raisebox{-\ht\treeboxtwo}{\usebox{\treeboxtwo}}}%
\hspace*{\daughteroffsetone}%
\raisebox{-\ht\treeboxone}{\usebox{\treeboxone}}%
\end{flushleft}%
\end{minipage}}%
\setlength{\treeoffsetone}{\parentoffset}%
\addtolength{\treeoffsetone}{0.5\wd\parentbox}%
\setlength{\treeshiftone}{0pt}%
\setlength{\treewidthone}{\treewidth}%
\sbox{\treeboxone}{\usebox{\treebox}}\poptree\poptree\poptree\poptree%
\else\typeout{QobiTeX warning--- Can't handle #1 branching}\fi\fi\fi\fi\fi}
\newcommand{\faketreewidth}[1]{%
\sbox{\parentbox}{\ontop{#1}}%
\setlength{\treewidthone}{0.5\wd\parentbox}%
\addtolength{\treewidthone}{\treeoffsetone}%
\setlength{\treeshiftone}{\treeoffsetone}%
\addtolength{\treeshiftone}{-0.5\wd\parentbox}}
\newcommand{\tree}{%
\usebox{\treeboxone}
\setlength{\treeoffsetone}{\treeoffsettwo}%
\sbox{\treeboxone}{\usebox{\treeboxtwo}}%
\poptree}

\newcommand{\derives}{\stackrel{\ast}{\Rightarrow}}

\documentstyle[acl]{article}

\title{Efficient Algorithms for Parsing the DOP Model
\thanks{\hspace{1em}I would like to acknowledge support from National
Science Foundation Grant IRI-9350192 and a National Science Foundation
Graduate Student Fellowship.  I would also like to thank Rens Bod,
Stan Chen, Andrew Kehler, David Magerman, Wheeler Ruml, Stuart
Shieber, and Khalil Sima'an for helpful discussions, and comments on
earlier drafts, and the comments of the anonymous reviewers.} }

\author{Joshua Goodman\\
        Harvard University\\
	33 Oxford St. \\
        Cambridge, MA 02138\\
        goodman@das.harvard.edu}

\begin{document}
\bibliographystyle{acl}
\maketitle

\begin{abstract}
Excellent results have been reported for Data-Oriented Parsing (DOP)
of natural language texts \cite{Bod:93a}.  Unfortunately, existing
algorithms are both computationally intensive and difficult to
implement.  Previous algorithms are expensive due to two factors: the
exponential number of rules that must be generated and the use of a
Monte Carlo parsing algorithm.  In this paper we solve the first
problem by a novel reduction of the DOP model to a small, equivalent
probabilistic context-free grammar.  We solve the second problem by a
novel deterministic parsing strategy that maximizes the expected
number of correct constituents, rather than the probability of a
correct parse tree.  Using the optimizations, experiments yield a 97\%
crossing brackets rate and 88\% zero crossing brackets rate.  This
differs significantly from the results reported by Bod, and is
comparable to results from a duplication of Pereira and Schabes's
\shortcite{Pereira:92a} experiment on the same data.  We show that Bod's
results are at least partially due to an extremely fortuitous choice
of test data, and partially due to using cleaner data than other
researchers.
\end{abstract}

\section{Introduction}
The Data-Oriented Parsing (DOP) model has a short, interesting, and
controversial history.  It was introduced by Remko Scha
\shortcite{Scha:90a}, and was then studied by Rens Bod.
Unfortunately, Bod \shortcite{Bod:93a,Bod:92a} was not able to find an
efficient exact algorithm for parsing using the model; however he did
discover and implement Monte Carlo approximations.  He tested these
algorithms on a cleaned up version of the ATIS corpus, and achieved
some very exciting results, reportedly getting 96\% of his test set
exactly correct, a huge improvement over previous results.  For
instance, Bod \shortcite{Bod:93b} compares these results to Schabes
\shortcite{Schabes:93a}, in which, for short sentences, 30\% of the
sentences have no crossing brackets (a much easier measure than exact
match).  Thus, Bod achieves an extraordinary 8-fold error rate
reduction.

Not surprisingly, other researchers attempted to duplicate these
results, but due to a lack of details of the parsing algorithm in his
publications, these other researchers were not able to confirm the
results (Magerman, Lafferty, personal communication).  Even Bod's
thesis \cite{Bod:95a} does not contain enough information to replicate
his results.

Parsing using the DOP model is especially difficult.  The model can be
summarized as a special kind of Stochastic Tree Substitution Grammar
(STSG): given a bracketed, labelled training corpus, let {\em
every} subtree of that corpus be an elementary tree, with a
probability proportional to the number of occurrences of that subtree
in the training corpus.  Unfortunately, the number of trees is in
general exponential in the size of the training corpus trees,
producing an unwieldy grammar.

In this paper, we introduce a reduction of the DOP model to an exactly
equivalent Probabilistic Context Free Grammar (PCFG) that is linear in
the number of nodes in the training data.  Next, we present an
algorithm for parsing, which returns the parse that is expected to
have the largest number of correct constituents.  We use the
reduction and algorithm to parse held out test data, comparing these
results to a replication of Pereira and Schabes \shortcite{Pereira:92a} on
the same data.  These results are disappointing: the PCFG
implementation of the DOP model performs about the same as the Pereira
and Schabes method.  We present an analysis of the runtime of our
algorithm and Bod's.  Finally, we analyze Bod's data, showing that
some of the difference between our performance and his is due to a
fortuitous choice of test data.

This paper contains the first published replication of the full DOP
model, i.e. using a parser which sums over derivations.  It also
contains algorithms implementing the model with significantly fewer
resources than previously needed.  Furthermore, for the first time,
the DOP model is compared on the same data to a competing model.

\section{Previous Research}
\label{sec:previous}
The DOP model itself is extremely simple and can be described as
follows: for every sentence in a parsed training corpus, extract every
subtree.  In general, the number of subtrees will be very large,
typically exponential in sentence length.  Now, use these trees to
form a Stochastic Tree Substitution Grammar (STSG).  There are two
ways to define a STSG: either as a Stochastic Tree Adjoining Grammar
\cite{Schabes:92a} restricted to substitution operations, or as an
extended PCFG in which entire trees may occur on the right hand side,
instead of just strings of terminals and non-terminals.

\begin{figure}
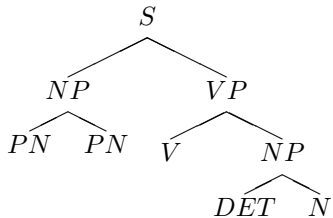


\begin{center}
\leaf{$PN$}
\leaf{$PN$}
\branch{2}{$NP$}
\leaf{$V$}
\leaf{$DET$}
\leaf{$N$}
\branch{2}{$NP$}
\branch{2}{$VP$}
\branch{2}{$S$}
\tree
\end{center}
\caption{Training corpus tree for DOP example}
\label{fig:maintree}
\end{figure}

\begin{figure*}
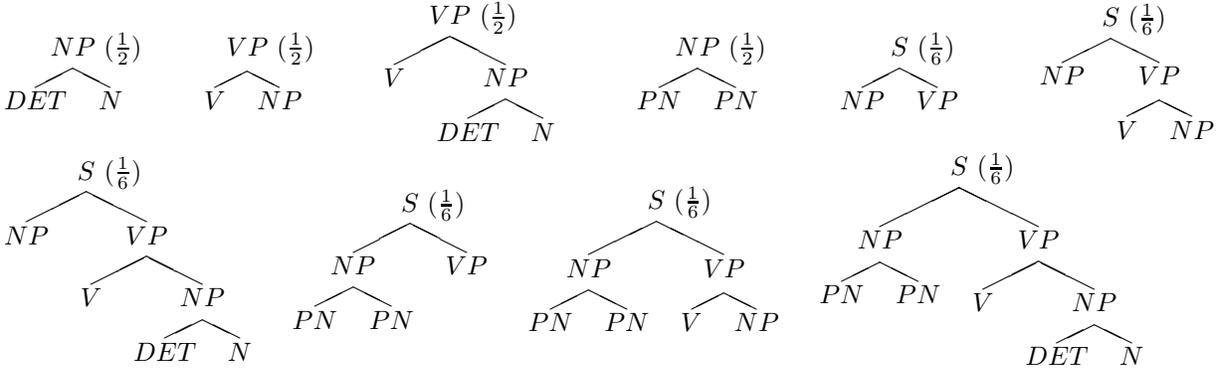

\leaf{$PN$}
\leaf{$PN$}
\branch{2}{$NP$}
\leaf{$V$}
\leaf{$DET$}
\leaf{$N$}
\branch{2}{$NP$}
\branch{2}{$VP$}
\branch{2}{$S$\makebox[0in][l]{ $(\frac{1}{6})$}}

\leaf{$PN$}
\leaf{$PN$}
\branch{2}{$NP$}
\leaf{$V$}
\leaf{$NP$}
\branch{2}{$VP$}
\branch{2}{$S$\makebox[0in][l]{ $(\frac{1}{6})$}}

\leaf{$PN$}
\leaf{$PN$}
\branch{2}{$NP$}
\leaf{$VP$}
\branch{2}{$S$\makebox[0in][l]{ $(\frac{1}{6})$}}

\leaf{$NP$}
\leaf{$V$}
\leaf{$DET$}
\leaf{$N$}
\branch{2}{$NP$}
\branch{2}{$VP$}
\branch{2}{$S$\makebox[0in][l]{ $(\frac{1}{6})$}}

\leaf{$NP$}
\leaf{$V$}
\leaf{$NP$}
\branch{2}{$VP$}
\branch{2}{$S$\makebox[0in][l]{ $(\frac{1}{6})$}}

\leaf{$NP$}
\leaf{$VP$}
\branch{2}{$S$\makebox[0in][l]{ $(\frac{1}{6})$}}

\leaf{$PN$}
\leaf{$PN$}
\branch{2}{$NP$\makebox[0in][l]{ $(\frac{1}{2})$}}

\leaf{$V$}
\leaf{$DET$}
\leaf{$N$}
\branch{2}{$NP$}
\branch{2}{$VP$\makebox[0in][l]{ $(\frac{1}{2})$}}

\leaf{$V$}
\leaf{$NP$}
\branch{2}{$VP$\makebox[0in][l]{ $(\frac{1}{2})$}}

\leaf{$DET$}
\leaf{$N$}
\branch{2}{$NP$\makebox[0in][l]{ $(\frac{1}{2})$}}

\tree \tree \tree \tree \tree \tree \tree \tree \tree \tree 
\caption{Sample STSG Produced from DOP Model}
\label{fig:sampleDOP}
\end{figure*}

Given the tree of Figure \ref{fig:maintree}, we can use the DOP model
to convert it into the STSG of Figure \ref{fig:sampleDOP}.  The
numbers in parentheses represent the probabilities.  These trees can
be combined in various ways to parse sentences.

In theory, the DOP model has several advantages over other models.
Unlike a PCFG, the use of trees allows capturing large contexts,
making the model more sensitive.  Since every subtree is included,
even trivial ones corresponding to rules in a PCFG, novel sentences
with unseen contexts can still be parsed.

\begin{figure}
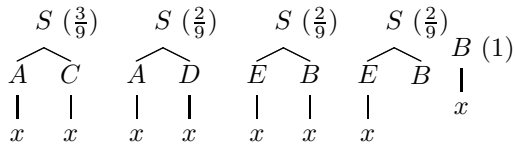

\begin{center}
\leaf{$x$}
\branch{1}{$A$}
\leaf{$x$}
\branch{1}{$C$} \faketreewidth{$D$}
\branch{2}{$S \makebox[0in][l]{ $(\frac{3}{9})$}$}
\tree
\hspace{-0.25in}
\leaf{$x$}
\branch{1}{$A$}
\leaf{$x$}
\branch{1}{$D$}
\branch{2}{$S \makebox[0in][l]{ $(\frac{2}{9})$}$}
\tree
\hspace{-0.25in}
\leaf{$x$}
\branch{1}{$E$}
\leaf{$x$}
\branch{1}{$B$}
\branch{2}{$S \makebox[0in][l]{ $(\frac{2}{9})$}$}
\tree
\hspace{-0.25in}
\leaf{$x$}
\branch{1}{$E$}
\leaf{$B$}
\branch{2}{$S \makebox[0in][l]{ $(\frac{2}{9})$}$}
\tree
\hspace{-0.25in}
\leaf{$x$}
\branch{1}{$B \makebox[0in][l]{ $(1)$}$}
\tree
\hspace{-0.25in}
\end{center}
\caption{Simple Example STSG}
\label{fig:simpleSTSG}
\end{figure}

Unfortunately, the number of subtrees is huge; therefore Bod randomly
samples 5\% of the subtrees, throwing away the rest.  This
significantly speeds up  parsing.

There are two existing ways to parse using the DOP model.  First, one
can find the most probable {\em derivation}.  That is, there can be
many ways a given sentence could be derived from the STSG.  Using the
most probable derivation criterion, one simply finds the most probable
way that a sentence could be produced.  Figure \ref{fig:simpleSTSG}
shows a simple example STSG.  For the string $x x$, what is the most
probable derivation?  The parse tree
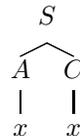
\begin{center}
\leaf{$x$}
\branch{1}{$A$}
\leaf{$x$}
\branch{1}{$C$}\faketreewidth{$D$}
\branch{2}{$S$}
\tree
\end{center}
\noindent has probability $\frac{3}{9}$ of being generated by the
trivial derivation containing a single tree.  This tree corresponds to
the most probable derivation of $x x$.

One could try to find the most probable parse {\em tree}.  For a given
sentence and a given parse tree, there are many different derivations
that could lead to that parse tree.  The probability of the parse
tree is the sum of the probabilities of the derivations.  Given our
example, there are two different ways to generate the parse tree
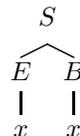
\begin{center}
\leaf{$x$}
\branch{1}{$E$}
\leaf{$x$}
\branch{1}{$B$}
\branch{2}{$S$}
\tree
\end{center}
\noindent each with probability $\frac{2}{9}$, so that the parse tree
has probability $\frac{4}{9}$.  This parse tree is most probable.

Bod \shortcite{Bod:93a} shows how to approximate this most probable parse
using a Monte Carlo algorithm.  The algorithm randomly samples
possible derivations, then finds the tree with the most sampled
derivations.  Bod shows that the most probable parse yields better
performance than the most probable derivation on the exact match
criterion.

Khalil Sima'an \shortcite{Sima'an:96a} implemented a version
of the DOP model, which parses efficiently by limiting the number of
trees used and by using an efficient most probable derivation model.
His experiments differed from ours and Bod's in many ways, including
his use of a different version of the ATIS corpus; the use of word
strings, rather than part of speech strings; and the fact that he did
not parse sentences containing unknown words, effectively throwing out
the most difficult sentences.  Furthermore, Sima'an limited the number
of substitution sites for his trees, effectively using a subset of the
DOP model.

\section{Reduction of DOP to PCFG}
\label{sec:reduction}

Unfortunately, Bod's reduction to a STSG is extremely expensive, even
when throwing away 95\% of the grammar.  Fortunately, it is possible
to find an equivalent PCFG that contains exactly eight PCFG rules for
each node in the training data; thus it is $O(n)$.  Because this
reduction is so much smaller, we do not discard any of the grammar
when using it.  The PCFG is equivalent in two senses: first it
generates the same strings with the same probabilities; second, using
an isomorphism defined below, it generates the same trees with the
same probabilities, although one must sum over several PCFG trees for
each STSG tree.

To show this reduction and equivalence, we must first define some
terminology.  We assign every node in every tree a unique number,
which we will call its address.  Let $A@k$ denote the node at address
$k$, where $A$ is the non-terminal labeling that node.  We will need
to create one new non-terminal for each node in the training data.  We
will call this non-terminal $A_k$.  We will call non-terminals of this
form ``interior'' non-terminals, and the original non-terminals in the
parse trees ``exterior''.

Let $a_j$ represent the number of subtrees headed by the node $A@j$.
Let $a$ represent the number of subtrees headed by nodes with
non-terminal $A$, that is $a=\sum_j a_j$.

Consider a node $A@j$ of the form:
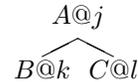
\begin{center}
\leaf{$B@k$} \faketreewidth{AA}
\leaf{$C@l$} \faketreewidth{AA}
\branch{2}{$A@j$}
\tree
\end{center}
How many subtrees does it have?  Consider first the possibilities on
the left branch.  There are $b_k$ non-trivial subtrees headed by
$B@k$, and there is also the trivial case where the left node is
simply $B$.  Thus there are $b_k+1$ different possibilities on the
left branch.  Similarly, for the right branch there are $c_l+1$
possibilities.  We can create a subtree by choosing any possible left
subtree and any possible right subtree.  Thus, there are
$a_j=(b_k+1)(c_l+1)$ possible subtrees headed by $A@j$.  In our
example tree of Figure \ref{fig:maintree}, both noun phrases have
exactly one subtree: $np_4=np_2=1$; the verb phrase has 2 subtrees:
$vp_3=2$; and the sentence has 6: $s_1=6$.  These numbers correspond to
the number of subtrees in Figure \ref{fig:sampleDOP}.

We will call a PCFG subderivation isomorphic to a STSG tree if the
subderivation begins with an external non-terminal, uses internal
non-terminals for intermediate steps, and ends with external
non-terminals.  For instance, consider the tree
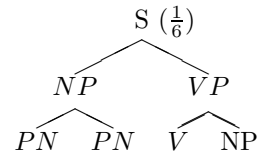
\begin{center}
\leaf{$PN$}
\leaf{$PN$}
\branch{2}{$NP$}
\leaf{$V$}
\leaf{NP}
\branch{2}{$VP$}
\branch{2}{S\makebox[0in][l]{ $(\frac{1}{6})$}}
\tree
\end{center}
taken from Figure \ref{fig:sampleDOP}.  The following PCFG subderivation
is isomorphic: $S \Rightarrow\ \:NP@1\ \:VP@2 \Rightarrow PN\ \:PN\ \:VP@2
\Rightarrow PN\ \:PN\ \:V\ \:NP$.  We say that a PCFG derivation is
isomorphic to a STSG derivation if there is a corresponding PCFG
subderivation for every step in the STSG derivation.

We will give a simple small PCFG with the following surprising
property: for every subtree in the training corpus headed by $A$, the
grammar will generate an isomorphic subderivation with probability
$1/a$.  In other words, rather than using the large, explicit STSG, we
can use this small PCFG that generates isomorphic derivations, with
identical probabilities.

The construction is as follows.  For a node such as
\begin{center}
\leaf{$B@k$} \faketreewidth{AA}
\leaf{$C@l$} \faketreewidth{AA}
\branch{2}{$A@j$}
\tree
\end{center}
we will generate the following eight
PCFG rules, where the number in parentheses following a rule is its
probability.
\begin{equation}
\begin{array}{llll}
A_j \rightarrow B C & (1/a_j) &
A \rightarrow B C & (1/a) \\
A_j \rightarrow B_k C & (b_k/a_j) &
A \rightarrow B_k C & (b_k/a) \\
A_j \rightarrow B C_l & (c_l/a_j) &
A \rightarrow B C_l & (c_l/a) \\
A_j \rightarrow B_k C_l & (b_k c_l/a_j) &
A \rightarrow B_k C_l & (b_k c_l/a) \\
\end{array}
\end{equation}  

We will show that sub\-der\-iva\-tions head\-ed by $A$ with ex\-tern\-al
non-terminals at the roots and leaves, internal non-terminals
elsewhere have probability $1/a$.  Subderivations headed by $A_j$ with
external non-terminals only at the leaves, internal non-terminals
elsewhere, have probability $1/a_j$.  The proof is by
induction on the depth of the trees.

For trees of depth 1, there are two cases:
\begin{center}
\leaf{$B$} \faketreewidth{AA}
\leaf{$C$} \faketreewidth{AA}
\branch{2}{$A$}
\tree
\leaf{$B$} \faketreewidth{AA}
\leaf{$C$} \faketreewidth{AA}
\branch{2}{$A@j$}
\tree
\end{center}
Trivially, these trees have the required probabilities.

Now, assume that the theorem is true for trees of depth $n$ or less.
We show that it holds for trees of depth $n+1$.  There are eight
cases, one for each of the eight rules.  We show two of them.  Let
$\begin{array}{c} \tiny B@k\\ \tiny \vdots\end{array}$ represent a
tree of at most depth $n$ with external leaves, headed by $B@k$, and
with internal intermediate non-terminals.  Then, for trees such as
\begin{center}
\leaf{$\begin{array}{c}B@k\\ \vdots\end{array}$} \faketreewidth{AA}
\leaf{$\begin{array}{c}C@l\\ \vdots\end{array}$} \faketreewidth{AA}
\branch{2}{$A@j$}
\tree
\end{center}
the probability of the tree is $\frac{1}{b_k}\frac{1}{c_l}\frac{b_k
c_l}{a_j}=\frac{1}{a_j}$.  Similarly, for another case, trees headed by
\begin{center}
\leaf{$B@k$} \faketreewidth{AA}
\leaf{$C$} \faketreewidth{AA}
\branch{2}{$A$}
\tree
\end{center}
the probability of the tree is
$\frac{1}{b_k}\frac{b_k}{a}=\frac{1}{a}$.  The other six cases follow
trivially with similar reasoning.

\begin{figure}
\begin{center}
\begin{tabular}{c}
PCFG derivation \\
4 productions \\
\leaf{$PN$}
\leaf{$PN$}
\branch{2}{$NP@3$}
\leaf{$V$}
\leaf{$DET$}
\leaf{$N$}
\branch{2}{$NP$}
\branch{2}{$VP@1$}
\branch{2}{$S$}
$\tree$ \\

\medskip\\
STSG derivation \\
2 subtrees \\
\leaf{$PN$}
\leaf{$PN$}
\branch{2}{$NP$}
\leaf{$V$}
\leaf{$DET$}
\leaf{$N$}
\branch{2}{$\begin{array}{cc}NP\\ \vspace{0in} \\ NP\end{array}$}
\branch{2}{$VP$}
\branch{2}{$S$}
$\tree$ \\

\end{tabular}
\end{center}
\caption{Example of Isomorphic Derivation}
\label{fig:isomorphic}
\end{figure}

We call a PCFG derivation isomorphic to a STSG derivation if for every
substitution in the STSG there is a corresponding subderivation in the
PCFG.  Figure \ref{fig:isomorphic} contains an example of isomorphic
derivations, using two subtrees in the STSG and four productions in
the PCFG.

We call a PCFG tree isomorphic to a STSG tree if they are identical
when internal non-terminals are changed to external non-terminals.
Our main theorem is that this construction produces PCFG trees
isomorphic to the STSG trees with equal probability.  If every subtree
in the training corpus occurred exactly once, this would be trivial to
prove.  For every STSG subderivation, there would be an isomorphic
PCFG subderivation, with equal probability.  Thus for every STSG
derivation, there would be an isomorphic PCFG derivation, with equal
probability.  Thus every STSG tree would be produced by the PCFG with
equal probability.

However, it is extremely likely that some subtrees, especially
trivial ones like
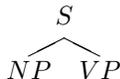
\begin{center}
\leaf{$NP$} \faketreewidth{AA}
\leaf{$VP$} \faketreewidth{AA}
\branch{2}{$S$}
\tree
\end{center}
will occur repeatedly.    

If the STSG formalism were modified slightly, so that trees could
occur multiple times, then our relationship could be made one to one.
Consider a modified form of the DOP model, in which when subtrees
occurred multiple times in the training corpus, their counts were not
merged: both identical trees are added to the grammar.  Each of these
trees will have a lower probability than if their counts were merged.
This would change the probabilities of the derivations; however the
probabilities of parse trees would not change, since there would be
correspondingly more derivations for each tree.  Now, the desired one
to one relationship holds: for every derivation in the new STSG there
is an isomorphic derivation in the PCFG with equal probability.  Thus,
summing over all derivations of a tree in the STSG yields the same
probability as summing over all the isomorphic derivations in the
PCFG.  Thus, every STSG tree would be produced by the PCFG with equal
probability.

It follows trivially from this that no extra trees are produced by the
PCFG.  Since the total probability of the trees produced by the STSG
is 1, and the PCFG produces these trees with the same probability, no
probability is ``left over'' for any other trees.

\section{Parsing Algorithm}
\label{sec:algorithm}
There are several different evaluation metrics one could use for
finding the best parse.  In the section covering previous research, we
considered the most probable derivation and the most probable parse
tree.  There is one more metric we could consider.  If our performance
evaluation were based on the number of constituents correct, using
measures similar to the crossing brackets measure, we would want the
parse tree that was most likely to have the largest number of correct
constituents.  With this criterion and the example grammar of Figure
\ref{fig:simpleSTSG}, the best parse tree would be
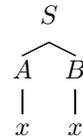
\begin{center}
\leaf{$x$}
\branch{1}{$A$}
\leaf{$x$}
\branch{1}{$B$}\faketreewidth{$D$}
\branch{2}{$S$}
\tree
\end{center}
The probability that the $S$ constituent is correct is $1.0$, while
the probability that the $A$ constituent is correct is $\frac{5}{9}$,
and the probability that the $B$ constituent is correct is
$\frac{4}{9}$.  Thus, this tree has on average $2$ constituents
correct.  All other trees will have fewer constituents correct on
average.  We call the best parse tree under this criterion the {\em
Maximum Constituents Parse}.  Notice that this parse tree cannot even
be produced by the grammar: each of its constituents is good,
but it is not necessarily good when considered as a full tree.

Bod \shortcite{Bod:93c, Bod:95a} shows that the most probable
derivation does not perform as well as the most probable parse for the
DOP model, getting 65\% exact match for the most probable derivation,
versus 96\% correct for the most probable parse.  This is not
surprising, since each parse tree can be derived by many different
derivations; the most probable parse criterion takes all possible
derivations into account.  Similarly, the Maximum Constituents Parse
is also derived from the sum of many different derivations.
Furthermore, although the Maximum Constituents Parse should not do as
well on the exact match criterion, it should perform even better on
the percent constituents correct criterion.  We have previously
performed a detailed comparison between the most likely parse, and the
Maximum Constituents Parse for Probabilistic Context Free Grammars
\cite{Goodman:96a}; we showed that the two have very similar
performance on a broad range of measures, with at most a 10\%
difference in error rate (i.e., a change from 10\% error rate to 9\%
error rate.)  We therefore think that it is reasonable to use a
Maximum Constituents Parser to parse the DOP model.

The pars\-ing al\-gor\-ithm is a var\-i\-ation on the
In\-side-Out\-side al\-gor\-ithm, developed by Baker \shortcite{Baker:79b}
and discussed in detail by Lari and Young \shortcite{Lari:90a}.  However,
while the Inside-Outside algorithm is a grammar re-estimation
algorithm, the algorithm presented here is just a parsing algorithm.
It is closely related to a similar algorithm used for Hidden Markov
Models \cite{Rabiner:89a} for finding the most likely state at each
time.  However, unlike in the HMM case where the algorithm produces a
simple state sequence, in the PCFG case a parse tree is produced,
resulting in additional constraints.

A formal derivation of a very similar algorithm is given elsewhere
\cite{Goodman:96a}; only the intuition is given here.  The algorithm
can be summarized as follows.  First, for each potential constituent,
where a constituent is a non-terminal, a start position, and an end
position, find the probability that that constituent is in the parse.
After that, put the most likely constituents together to form a parse
tree, using dynamic programming.

The probability that a potential constituent occurs in the correct
parse tree, $P(X \derives w_s ... w_t | S \derives w_1 ... w_n)$, will
be called $g(s, t, X)$.  In words, it is the probability that, given
the sentence $w_1...w_n$, a symbol X generates $w_s...w_t$.  We can
compute this probability using elements of the Inside-Outside
algorithm.  First, compute the inside probabilities, $e(s, t, X) = P(X
\derives w_s ... w_t)$.  Second, compute the outside probabilities,
$f(s, t, X) = P(S \derives w_1 ... w_{s-1} X w_{t+1} ... w_n)$.
Third, compute the matrix $g(s, t, X)$:

{
\begin{eqnarray*}
\lefteqn{g(s, t, X)} \\
& = & \frac{P(S \derives w_1...w_{s-1} X w_{t+1}...w_n )
                        P(X \derives w_s...w_{t}) }
                      { \small P(S \derives w_1...w_n) } \\
           & = & f(s, t, X) \times e(s, t, X) / e(1, n, S)
\end{eqnarray*}
}

\begin{figure}
\begin{verbatim}
for length := 2 to n
   for s := 1 to n-length+1
      t := s + length - 1;
      for all non-terminals X
         sum[X] := g(s, t, X);
      loop over addresses k
         let X := non-terminal at k;
         let sum[X] := sum[X] + g(s,t,X_k);
      loop over non-terminals X
         let max_X := arg max of sum[X]
      loop over r such that s <= r < t
         let best_split := 
            max of maxc[s,r] + maxc[r+1,t];
      maxc[s,t] := sum[max_X] + best_split;
\end{verbatim}
\caption{Maximum Constituents Data-Oriented Parsing Algorithm}
\label{fig:maxcons}
\end{figure}

Once the matrix $g(s, t, X)$ is computed, a dynamic programming
algorithm can be used to determine the best parse, in the sense of
maximizing the number of constituents expected correct.
Figure \ref{fig:maxcons} shows pseudocode for a simplified form of
this algorithm.

For a grammar with $g$ nonterminals and training data of size $T$, the
run time of the algorithm is $O(Tn^2 + gn^3 + n^3)$ since there are
two layers of outer loops, each with run time at most $n$, and inner
loops, over addresses (training data), nonterminals and $n$.  However,
this is dominated by the computation of the Inside and Outside
probabilities, which takes time $O(rn^3)$, for a grammar with $r$
rules.  Since there are eight rules for every node in the training
data, this is $O(Tn^3)$.

By modifying the algorithm slightly to record the actual split used at each
node, we can recover the best parse.  The entry {\tt maxc[1, n]}
contains the expected number of correct constituents, given the model.

\section{Experimental Results and Discussion}
\label{sec:results}
\begin{table*}
\begin{center}
\begin {tabular}{|l||r|r|r|r|r|} \hline
       Criteria & Min    & Max    & Range  & Mean  & StdDev  \\ \hline
Cross Brack DOP &86.53\% &96.06\% & 9.53\% &90.15\% & 2.65\%  \\ \hline
Cross Brack P\&S &86.99\% &94.41\% & 7.42\% &90.18\% & 2.59\%  \\ \hline
Cross Brack DOP$-$P\&S &-3.79\% & 2.87\% & 6.66\% &-0.03\% & 2.34\%  \\ \hline
Zero Cross Brack DOP &60.23\% &75.86\% &15.63\% &66.11\% & 5.56\%  \\ \hline
Zero Cross Brack P\&S &54.02\% &78.16\% &24.14\% &63.94\% & 7.34\%  \\ \hline
Zero Cross Brack DOP$-$P\&S &-5.68\% &11.36\% &17.05\% & 2.17\% & 5.57\%  \\ \hline
\end{tabular}
\end{center}
\caption{DOP versus Pereira and Schabes on Minimally Edited ATIS}
\label{tab:results}
\end{table*}

\begin{table*}
\begin{center}
\begin {tabular}{|l||r|r|r|r|r|} \hline
       Criteria & Min    & Max    & Range  & Mean  & StdDev  \\ \hline 
Cross Brack DOP &95.63\% &98.62\% & 2.99\% &97.16\% & 0.93\%  \\ \hline 
Cross Brack P\&S &94.08\% &97.87\% & 3.79\% &96.11\% & 1.14\%  \\ \hline 
Cross Brack DOP$-$P\&S &-0.16\% & 3.03\% & 3.19\% & 1.05\% & 1.04\%  \\ \hline 
Zero Cross Brack DOP &78.67\% &90.67\% &12.00\% &86.13\% & 3.99\%  \\ \hline 
Zero Cross Brack P\&S &70.67\% &88.00\% &17.33\% &79.20\% & 5.97\%  \\ \hline 
Zero Cross Brack DOP$-$P\&S &-1.33\% &20.00\% &21.33\% & 6.93\% & 5.65\%  \\ \hline 
Exact Match DOP &58.67\% &68.00\% & 9.33\% &63.33\% & 3.22\%  \\ \hline 
\end{tabular}
\end{center}
\caption{DOP versus Pereira and Schabes on Bod's Data}
\label{tab:Bodresults}
\end{table*}

We are grateful to Bod for supplying the data that he used for
his experiments \cite{Bod:95b, Bod:95a, Bod:93a}.  The original ATIS
data from the Penn Tree Bank, version 0.5, is very noisy; it is
difficult to even automatically read this data, due to inconsistencies
between files.  Researchers are thus left with the difficult decision
as to how to clean the data.  For this paper, we conducted two sets of
experiments: one using a minimally cleaned set of data,\footnote{A diff
file between the original ATIS data and the cleaned up version, in a
form usable by the ``ed'' program, is available by anonymous FTP from
ftp://ftp.das.harvard.edu/pub/goodman/atis-ed/ ti\_tb.par-ed and
ti\_tb.pos-ed.  Note that the number of changes made was small.  The
diff files sum to 457 bytes, versus 269,339 bytes for the original
files, or less than 0.2\%.}  
making our results comparable to previous
results; the other using the ATIS data prepared by Bod, which
contained much more significant revisions.

\begin{sloppypar}
Ten data sets were con\-struct\-ed by ran\-dom\-ly split\-ting
min\-i\-mal\-ly edit\-ed ATIS (Hemphill et al., 1990) \nocite{Hemphill:90a} sen\-ten\-ces in\-to a 700
sen\-t\-en\-ce train\-ing set, and 88 sentence test set, then discarding
sentences of length $> 30$.  For each of the ten sets, both the DOP
algorithm outlined here and the grammar induction experiment of
Pereira and Schabes were run.  Crossing brackets, zero crossing
brackets, and the paired differences are presented in Table
\ref{tab:results}.  All sentences output by the parser were made
binary branching (see the section covering analysis of Bod's data),
since otherwise the crossing brackets measures are meaningless
\cite{Magerman:94a}.  A few sentences were not parsable; these were
assigned right branching period high structure, a good heuristic
\cite{Brill:93a}.
\end{sloppypar}

We also ran experiments using Bod's data, 75 sentence test sets, and
no limit on sentence length.  However, while Bod provided us
with his data, he did not provide us with the split into test and
training that he used; as before we used ten random splits.  The
results are disappointing, as shown in Table \ref{tab:Bodresults}.
They are noticeably worse than those of Bod, and again very comparable
to those of Pereira and Schabes.  Whereas Bod reported 96\% exact
match, we got only 86\% using the less restrictive zero crossing
brackets criterion.  It is not clear what exactly accounts for these
differences.\footnote{Ideally, we would exactly reproduce these
experiments using Bod's algorithm.  Unfortunately, it was not possible
to get a full specification of the algorithm.}
It is also noteworthy that the results are much better on Bod's data
than on the minimally edited data: crossing brackets rates of 96\% and
97\% on Bod's data versus 90\% on minimally edited data.  Thus it
appears that part of Bod's extraordinary performance can be explained
by the fact that his data is much cleaner than the data used by other
researchers.

DOP does do slightly better on most measures.  We performed a
statistical analysis using a $t$-test on the paired differences
between DOP and Pereira and Schabes performance on each run.  On the
minimally edited ATIS data, the differences were statistically
insignificant, while on Bod's data the differences were statistically
significant beyond the 98'th percentile.  Our technique for finding
statistical significance is more strenuous than most: we assume that
since all test sentences were parsed with the same training data, all
results of a single run are correlated.  Thus we compare paired
differences of entire runs, rather than of sentences or constituents.
This makes it harder to achieve statistical significance.

Notice also the minimum and maximum columns of the ``DOP$-$P\&S'' lines,
constructed by finding for each of the paired runs the difference
between the DOP and the Pereira and Schabes algorithms.  Notice that
the minimum is usually negative, and the maximum is usually positive,
meaning that on some tests DOP did worse than Pereira and Schabes and
on some it did better.  It is important to run multiple tests,
especially with small test sets like these, in order to avoid
misleading results.

\section{Timing Analysis}
\label{sec:timing}

In this section, we examine the empirical runtime of our algorithm,
and analyze Bod's.
We also note that Bod's algorithm
will probably be particularly inefficient on longer sentences.

It takes about 6 seconds per sentence to run our algorithm on an HP
9000/715, versus 3.5 hours  to run Bod's algorithm on a Sparc 2
\cite{Bod:95b}.  Factoring in that the HP is roughly four
times faster than the Sparc,
the new algorithm is about 500 times faster.  Of course, some of this
difference may be due to differences in implementation, so this
estimate is fairly rough.


Furthermore, we believe Bod's analysis of his parsing algorithm is
flawed.  Letting $G$ represent grammar size, and $\epsilon$ represent
maximum estimation error, Bod correctly analyzes his runtime as
$O(Gn^3\epsilon^{-2})$.  However, Bod then neglects analysis of this
$\epsilon^{-2}$ term, assuming that it is constant.  Thus he concludes
that his algorithm runs in polynomial time.  However, for his
algorithm to have some reasonable chance of finding the most probable
parse, the number of times he must sample his data is at least
inversely proportional to the conditional probability of that parse.
For instance, if the maximum probability parse had probability $1/50$,
then he would need to sample at least $50$ times to be reasonably sure
of finding that parse.


Now, we note that the conditional probability of the most probable
parse tree will in general decline exponentially with sentence length.
We assume that the number of ambiguities in a sentence will increase
linearly with sentence length; if a five word sentence has on average
one ambiguity, then a ten word sentence will have two, etc.  A linear
increase in ambiguity will lead to an exponential decrease in
probability of the most probable parse.

Since the probability of the most probable parse decreases
exponentially in sentence length, the number of random samples needed
to find this most probable parse increases exponentially in sentence
length.  Thus, when using the Monte Carlo algorithm, one is left with
the uncomfortable choice of exponentially decreasing the probability
of finding the most probable parse, or exponentially increasing the
runtime.

We admit that this is a somewhat informal argument.  Still, the Monte
Carlo algorithm has never been tested on sentences longer than those
in the ATIS corpus; there is good reason to believe the algorithm will
not work as well on longer sentences.  Note that our algorithm has
true runtime $O(Tn^3)$, as shown previously.

\section{Analysis of Bod's Data}
\label{sec:analysis}

In the DOP model, a sentence cannot be given an exactly correct parse
unless all productions in the correct parse occur in the training set.
Thus, we can get an upper bound on performance by examining the test
corpus and finding which parse trees could not be generated using only
productions in the training corpus.  Unfortunately, while Bod provided
us with his data, he did not specify which sentences were test and
which were training.  We can however find an upper bound on average
case performance, as well as an upper bound on the probability that
any particular level of performance could be achieved.

\begin{table*}
\begin{center}
\begin{tabular}{|c|c|c|c|}
\hline
Original & Correct & Continued & Simple \\  \hline
\leaf{$B$}
\leaf{$C$}
\leaf{$D$}
\leaf{$E$}
\branch{4}{$A$}
\tree
&
\leaf{$B$}
\leaf{$C$}
\leaf{$D$}
\leaf{$E$}
\branch{2}{$*\_DE$}
\branch{2}{$*\_CDE$}
\branch{2}{$A$}
\tree
&
\leaf{$B$}
\leaf{$C$}
\leaf{$D$}
\leaf{$E$}
\branch{2}{$A\_*$}
\branch{2}{$A\_*$}
\branch{2}{$A$}
\tree
&
\leaf{$B$}
\leaf{$C$}
\leaf{$D$}
\leaf{$E$}
\branch{2}{$A$}
\branch{2}{$A$}
\branch{2}{$A$}
\tree
\\ \hline
\end{tabular}
\end{center}
\caption{Transformations from $N$-ary to Binary Branching Structures} 
\label{tab:convert}
\end{table*}

Bod randomly split his corpus into test and training.  According to
his thesis \cite[page 64]{Bod:95a}, only one of his 75 test sentences
had a correct parse which could not be generated from the training
data.  This turns out to be very surprising.  An analysis of Bod's
data shows that at least some of the difference in performance between
his results and ours must be due to an extraordinarily fortuitous
choice of test data.  It would be very interesting to see how our
algorithm performed on Bod's split into test and training, but he has
not provided us with this split.  Bod did examine versions of DOP that
smoothed, allowing productions which did not occur in the training
set; however his reference to coverage is with respect to a version
which does no smoothing.

\begin{table*}
\begin{center}
\begin{tabular}{|l||lr|lr|lr|}
\hline &
\multicolumn{2}{c|}{Correct}&
\multicolumn{2}{c|}{Continued}&
\multicolumn{2}{c|}{Simple}
\\ \hline
no unary & 0.78 & 0.0000002  &  0.88 & 0.0009484  &  0.90 & 0.0041096\\  \hline
unary    & 0.80 & 0.0000011  &  0.90 & 0.0037355  &  0.92 & 0.0150226\\  \hline
\end{tabular}
\end{center}
\caption{Probabilities of Sentences with Unique Productions/Test Data
with Ungeneratable Sentences}
\label{tab:under}
\end{table*}

In order to perform our analysis, we must determine certain details of
Bod's parser which affect the probability of having most sentences
correctly parsable.  When using a chart parser, as Bod did, three
problematic cases must be handled: $\epsilon$ productions, unary
productions, and $n$-ary ($n > 2$) productions.  The first two kinds
of productions can be handled with a probabilistic chart parser, but
large and difficult matrix manipulations are required
\cite{Stolcke:93a}; these manipulations would be especially difficult
given the size of Bod's grammar.  Examining Bod's data, we find he
removed $\epsilon$ productions.  We also assume that Bod made the same
choice we did and eliminated unary productions, given the difficulty
of correctly parsing them.  Bod himself does not know which technique
he used for $n$-ary productions, since the chart parser he used was
written by a third party (Bod, personal communication).
%
%

The $n$-ary productions can be parsed in a straightforward manner, by
converting them to binary branching form; however, there are at least
three different ways to convert them, as illustrated in Table
\ref{tab:convert}.  In method ``Correct'', the $n$-ary branching
productions are converted in such a way that no overgeneration is
introduced.  A set of special non-terminals is added, one for each
partial right hand side.  In method ``Continued'', a single new
non-terminal is introduced for each original non-terminal.  Because
these non-terminals occur in multiple contexts, some overgeneration is
introduced.  However, this overgeneration is constrained, so that
elements that tend to occur only at the beginning, middle, or end of
the right hand side of a production cannot occur somewhere else.  If
the ``Simple'' method is used, then no new non-terminals are
introduced; using this method, it is not possible to recover the
$n$-ary branching structure from the resulting parse tree, and
significant overgeneration occurs.

Table \ref{tab:under} shows the undergeneration probabilities for each
of these possible techniques for handling unary productions and
$n$-ary productions.\footnote{A perl script for analyzing Bod's data
is available by anon\-y\-mous FTP from
ftp://ftp.das.harvard.edu/pub/goodman/analyze.perl}
The first number in each column is the probability that
a sentence in the training data will have a production that occurs
nowhere else.  The second number is the probability that a test set of
75 sentences drawn from this database will have one ungeneratable
sentence: $75p^{74}(1-p)$.\footnote{Actually, this is a slight
overestimate for a few reasons, including the fact that the 75
sentences are drawn without replacement.  Also, consider a sentence
with a production that occurs only in one other sentence in the
corpus; there is some probability that both sentences will end up in
the test data, causing both to be ungeneratable.}

The table is arranged from least generous to most generous: in the
upper left hand corner is a technique Bod might reasonably have used;
in that case, the probability of getting the test set he described is
less than one in a million.  In the lower right corner we give Bod the
absolute maximum benefit of the doubt: we assume he used a parser
capable of parsing unary branching productions, that he used a very
overgenerating grammar, and that he used a loose definition of ``Exact
Match.''  Even in this case, there is only about a 1.5\% chance of
getting the test set Bod describes.

\section{Conclusion}

We have given efficient techniques for parsing the DOP model.  These
results are significant since the DOP model has perhaps the best
reported parsing accuracy; previously the full DOP model had not been
replicated due to the difficulty and computational complexity of the
existing algorithms.  We have also shown that previous results were
partially due to an unlikely choice of test data, and partially due to
the heavy cleaning of the data, which reduced the difficulty of the
task.

Of course, this research raises as many questions as it answers.  Were
previous results due only to the choice of test data, or are the
differences in implementation partly responsible?  In that case, there
is significant future work required to understand which differences
account for Bod's exceptional performance.  This will be complicated
by the fact that sufficient details of Bod's implementation are not
available.

This research also shows the importance of testing on more than one
small test set, as well as the importance of not making cross-corpus
comparisons; if a new corpus is required, then previous algorithms
should be duplicated for comparison.




\begin{thebibliography}{}

\bibitem[\protect\citename{Baker}1979]{Baker:79b}
J.K. Baker.
\newblock 1979.
\newblock Trainable grammars for speech recognition.
\newblock In {\em Proceedings of the Spring Conference of the Acoustical
  Society of America}, pages 547--550, Boston, MA, June.

\bibitem[\protect\citename{Bod}1992]{Bod:92a}
Rens Bod.
\newblock 1992.
\newblock Mathematical properties of the data oriented parsing model.
\newblock Paper presented at the {\em Third Meeting on Mathematics of Language
  (MOL3)}, Austin Texas.

\bibitem[\protect\citename{Bod}1993a]{Bod:93c}
Rens Bod.
\newblock 1993a.
\newblock Data-oriented parsing as a general framework for stochastic language
  processing.
\newblock In K.~Sikkel and A.~Nijholt, editors, {\em Parsing Natural Language}.
  Twente, The Netherlands.

\bibitem[\protect\citename{Bod}1993b]{Bod:93b}
Rens Bod.
\newblock 1993b.
\newblock Monte {C}arlo parsing.
\newblock In {\em Proceedings Third International Workshop on Parsing
  Technologies}, Tilburg/Durbury.

\bibitem[\protect\citename{Bod}1993c]{Bod:93a}
Rens Bod.
\newblock 1993c.
\newblock Using an annotated corpus as a stochastic grammar.
\newblock In {\em Proceedings of the Sixth Conference of the European Chapter
  of the ACL}, pages 37--44.

\bibitem[\protect\citename{Bod}1995a]{Bod:95a}
Rens Bod.
\newblock 1995a.
\newblock {\em Enriching Linguistics with Statistics: Performance Models of
  Natural Language}.
\newblock University of Amsterdam ILLC Dissertation Series 1995-14. Academische
  Pers, Amsterdam.

\bibitem[\protect\citename{Bod}1995b]{Bod:95b}
Rens Bod.
\newblock 1995b.
\newblock The problem of computing the most probable tree in data-oriented
  parsing and stochastic tree grammars.
\newblock In {\em Proceedings of the Seventh Conference of the European Chapter
  of the ACL}.

\bibitem[\protect\citename{Brill}1993]{Brill:93a}
Eric Brill.
\newblock 1993.
\newblock {\em A Corpus-Based Approach to Language Learning}.
\newblock {Ph.D.} thesis, University of Pennsylvania.

\bibitem[\protect\citename{Goodman}1996]{Goodman:96a}
Joshua Goodman.
\newblock 1996.
\newblock Parsing algorithms and metrics.
\newblock In {\em Proceedings of the 34th Annual Meeting of the ACL}.
\newblock To appear.

\bibitem[\protect\citename{Hemphill \bgroup et al.\egroup }1990]{Hemphill:90a}
Charles~T. Hemphill, John~J. Godfrey, and George~R. Doddington.
\newblock 1990.
\newblock The {ATIS} spoken language systems pilot corpus.
\newblock In {\em DARPA Speech and Natural Language Workshop}, Hidden Valley,
  Pennsylvania, June. Morgan Kaufmann.

\bibitem[\protect\citename{Lari and Young}1990]{Lari:90a}
K.~Lari and S.J. Young.
\newblock 1990.
\newblock The estimation of stochastic context-free grammars using the
  inside-outside algorithm.
\newblock {\em Computer Speech and Language}, 4:35--56.

\bibitem[\protect\citename{Magerman}1994]{Magerman:94a}
David Magerman.
\newblock 1994.
\newblock {\em Natural Language Parsing as Statistical Pattern Recognition}.
\newblock {Ph.D.} thesis, Stanford University University, February.

\bibitem[\protect\citename{Pereira and Schabes}1992]{Pereira:92a}
Fernando Pereira and Yves Schabes.
\newblock 1992.
\newblock {I}nside-{O}utside reestimation from partially bracketed corpora.
\newblock In {\em Proceedings of the 30th Annual Meeting of the ACL}, pages
  128--135, Newark, Delaware.

\bibitem[\protect\citename{Rabiner}1989]{Rabiner:89a}
L.R. Rabiner.
\newblock 1989.
\newblock A tutorial on hidden {M}arkov models and selected applications in
  speech recognition.
\newblock {\em Proceedings of the IEEE}, 77(2), February.

\bibitem[\protect\citename{Scha}1990]{Scha:90a}
R.~Scha.
\newblock 1990.
\newblock Language theory and language technology; competence and performance.
\newblock In Q.A.M. de~Kort and G.L.J. Leerdam, editors, {\em
  Computertoepassingen in de Neerlandistiek}. Landelijke Vereniging van
  Neerlandici (LVVN-jaarboek), Almere.
\newblock In Dutch.

\bibitem[\protect\citename{Schabes \bgroup et al.\egroup }1993]{Schabes:93a}
Yves Schabes, Michal Roth, and Randy Osborne.
\newblock 1993.
\newblock Parsing the {W}all {S}treet {J}ournal with the {I}nside-{O}utside
  algorithm.
\newblock In {\em Proceedings of the Sixth Conference of the European Chapter
  of the ACL}, pages 341--347.

\bibitem[\protect\citename{Schabes}1992]{Schabes:92a}
Y.~Schabes.
\newblock 1992.
\newblock Stochastic lexicalized tree-adjoining grammars.
\newblock In {\em Proceedings of the 14th International Conference on
  Computational Linguistics}.

\bibitem[\protect\citename{Sima'an}1996]{Sima'an:96a}
Khalil Sima'an.
\newblock 1996.
\newblock Efficient disambiguation by means of stochastic tree substitution
  grammars.
\newblock In R.~Mitkov and N.~Nicolov, editors, {\em Recent Advances in NLP
  1995}, volume 136 of {\em Current Issues in Linguistic Theory}. John
  Benjamins, Amsterdam.

\bibitem[\protect\citename{Stolcke}1993]{Stolcke:93a}
Andreas Stolcke.
\newblock 1993.
\newblock An efficient probabilistic context-free parsing algorithm that
  computes prefix probabilities.
\newblock Technical Report TR-93-065, International Computer Science Institute,
  Berkeley, CA.

\end{thebibliography}
\end{document}